\documentclass[conference]{IEEEtran}

\usepackage{cite}
\usepackage{times}
\usepackage{graphicx}
\usepackage{amsmath}
\usepackage{amssymb}
\usepackage{booktabs}
\usepackage{multirow}
\usepackage{array}
\usepackage{subcaption}
\usepackage{caption}
\usepackage[breaklinks=true,colorlinks,bookmarks=true]{hyperref}

\begin{document}

\title{\LARGE \bf
Automatic Vehicle Detection Using DETR: A Transformer-Based Approach for Navigating Treacherous Roads
}

\author{
\IEEEauthorblockN{Istiaq Ahmed Fahad}
\IEEEauthorblockA{Institute of Information Technology\\
University of Dhaka\\
Dhaka, Bangladesh\\
bsse1204@iit.du.ac.bd}
\and
\IEEEauthorblockN{Abdullah Ibne Hanif Arean}
\IEEEauthorblockA{Dept. of Computer Science and Engineering\\
University of Dhaka\\
Dhaka, Bangladesh\\
abdullaharean2613@gmail.com}
\and
\IEEEauthorblockN{Nazmus Sakib Ahmed}
\IEEEauthorblockA{Institute of Information Technology\\
University of Dhaka\\
Dhaka, Bangladesh\\
bsse1108@iit.du.ac.bd}
\and[\hfill\break\hbox{}\hfill]
\IEEEauthorblockN{Mahmudul Hasan}
\IEEEauthorblockA{Dept. of Computer Science and Engineering\\
University of Dhaka\\
Dhaka, Bangladesh\\
mahmudul.hhh@gmail.com}
}

\maketitle
\pagestyle{plain}
\thispagestyle{plain}

\begin{abstract}
    Automatic Vehicle Detection (AVD) in diverse driving environments presents unique challenges due to varying lighting conditions, road types, and vehicle types. Traditional methods, such as YOLO and Faster R-CNN, often struggle to cope with these complexities. As computer vision evolves, combining Convolutional Neural Networks (CNNs) with Transformer-based approaches offers promising opportunities for improving detection accuracy and efficiency. This study is the first to experiment with Detection Transformer (DETR) for automatic vehicle detection in complex and varied settings. We employ a Collaborative Hybrid Assignments Training scheme, Co-DETR, to enhance feature learning and attention mechanisms in DETR. By leveraging versatile label assignment strategies and introducing multiple parallel auxiliary heads, we provide more effective supervision during training and extract positive coordinates to boost training efficiency. Through extensive experiments on DETR variants and YOLO models, conducted using the BadODD dataset, we demonstrate the advantages of our approach. Our method achieves superior results and improved accuracy under diverse conditions, making it practical for real-world deployment. This work significantly advances autonomous navigation technology and opens new research avenues in object detection for autonomous vehicles. By integrating the strengths of CNNs and Transformers, we highlight the potential of DETR for robust and efficient vehicle detection in challenging driving environments.
\end{abstract}


\section{Introduction}

Autonomous navigation technology has witnessed remarkable advancements in recent years, revolutionizing various industries and promising safer and more efficient transportation systems. In particular, the deployment of Autonomous Vehicles (AVs) holds immense potential for transforming the way we commute and travel \cite{shalev2017formal, grigorescu2020survey}. However, the successful realization of fully autonomous vehicles requires robust object detection systems capable of accurately identifying and localizing objects in complex and diverse driving environments.

\begin{figure}[!htb]
        \centering
        \includegraphics[width=0.60\linewidth]{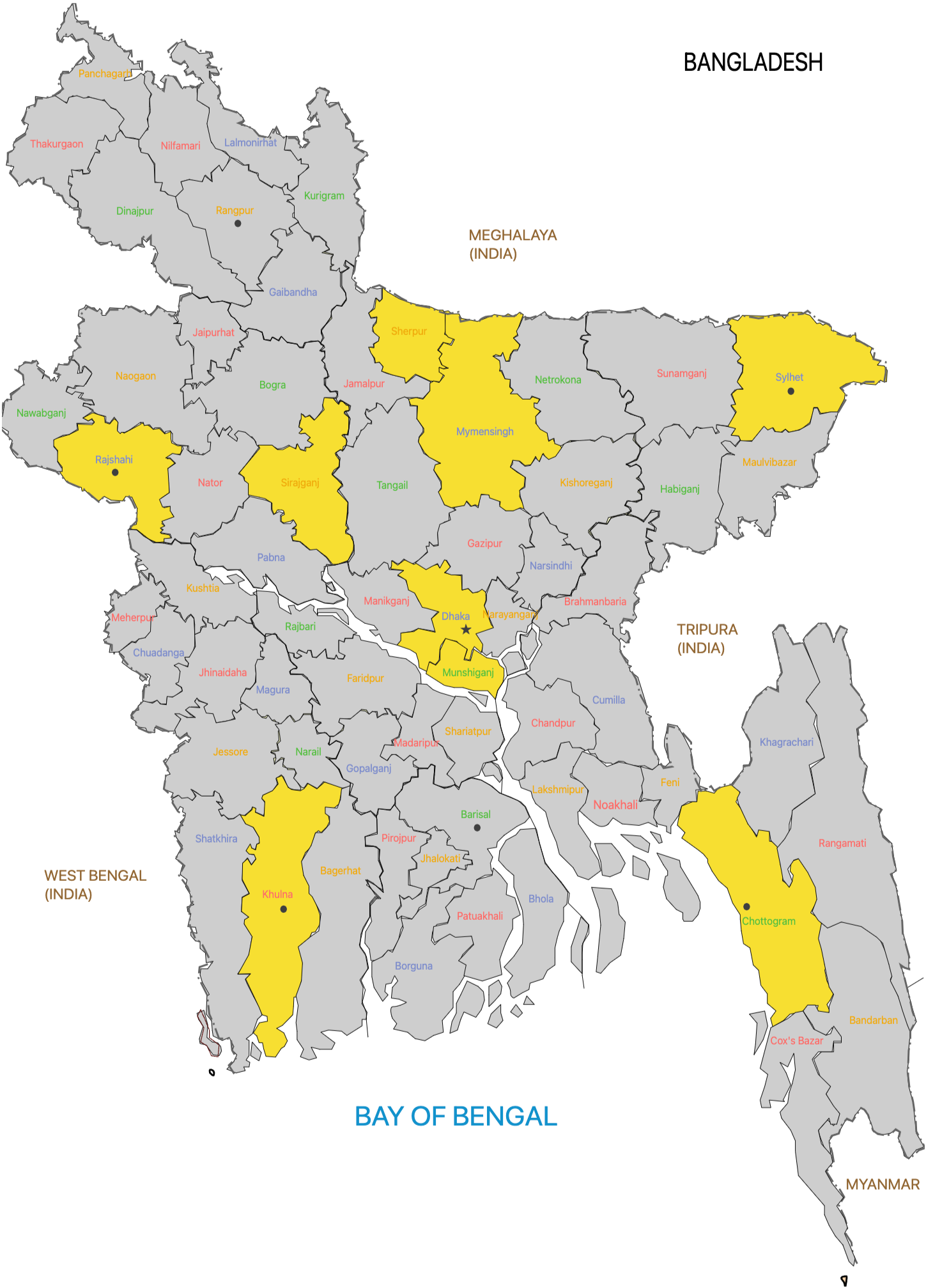}
        \caption{Districts of Bangladesh from which the BadODD data were collected}
        \label{fig:area-coverage}
\end{figure}

Object detection serves as a fundamental component in the perception stack of autonomous vehicles, enabling them to understand and interact with their surroundings effectively. Traditional object detection methods \cite{felzenszwalb2010object, viola2001rapid, dalal2005hog} often rely on handcrafted features and complex pipelines, which may struggle to generalize across different environments and exhibit limited scalability. Recent research has shifted toward end-to-end trainable deep learning approaches \cite{ren2015faster, redmon2016you, Carion2020DETR} to overcome these limitations, offering improved performance and adaptability.

This paper focuses on addressing the unique challenges of automatic vehicle detection in diverse driving environments. These environments are characterized by varying road conditions, diverse vehicle types, and challenging lighting conditions \cite{zhu2019visdrone, geiger2012kitti, caesar2020nuscenes}. To facilitate research and development in this domain, we used the BadODD dataset \cite{Baig2024BadODDBA}, a comprehensive dataset specifically curated for detecting autonomous driving objects in complex settings.

We fine-tuned DETR \cite{Carion2020DETR} with Collaborative Hybrid Assignments Training (Co-DETR) \cite{zong2023detrs} to enhance the performance of DETR-based object detectors. We evaluated this approach against YOLOv8m \cite{ultralytics2023yolov8} models, demonstrating improvements in efficiency and effectiveness through versatile label assignment strategies. This method addresses issues related to sparse supervision and inefficient feature learning, thereby improving the model’s ability to detect vehicles accurately in challenging scenarios. By combining the BadODD dataset with this training scheme, we aim to provide a robust framework for AVD in diverse driving conditions.

Our study pioneers the use of Detection Transformer (DETR) technology \cite{Carion2020DETR} for AVD in diverse driving environments. We fine-tuned the DETR model to enhance feature learning and attention mechanisms. Through experiments on the BadODD dataset, we demonstrate superior efficiency and enhanced accuracy compared to traditional methods like YOLO and Faster R-CNN \cite{ren2015faster}. This work validates DETR’s practical application in autonomous navigation and its potential to advance object detection in challenging driving scenarios.

\begin{itemize}
    \item Pioneers the use of the Detection Transformer (DETR) method for vehicle detection in diverse driving environments.
    \item Fine-tuned DETR with Collaborative Hybrid Assignments Training (Co-DETR) to enhance feature learning and attention mechanisms.
    \item Demonstrates thorough experiments on the BadODD dataset, showing superior efficiency and enhanced accuracy compared to traditional methods like YOLO and Faster R-CNN.
\end{itemize}

To outline our workflow, the paper is structured as follows: Section II reviews related work on autonomous vehicle detection. Section III provides a detailed description of the BadODD dataset, including coverage of Bangladesh districts, image and object statistics, and annotation details. Section IV elaborates on the methodology, including data preprocessing, model selection, and model specification. Section V outlines the experimental setup, detailing the hardware and software configurations used for training the models. Section VI presents the results and discussion, showcasing the performance of YOLOv8m and Co-DETR models. Section VII concludes the paper by summarizing the contributions and implications of the proposed approach for automatic vehicle detection in diverse driving environments. Hereafter, we use the term \textit{Co-DETR} to refer to \textit{DETRs with Collaborative Hybrid Assignments Training}.

    \begin{figure}[!htb]
        \centering
        \includegraphics[width=\linewidth]{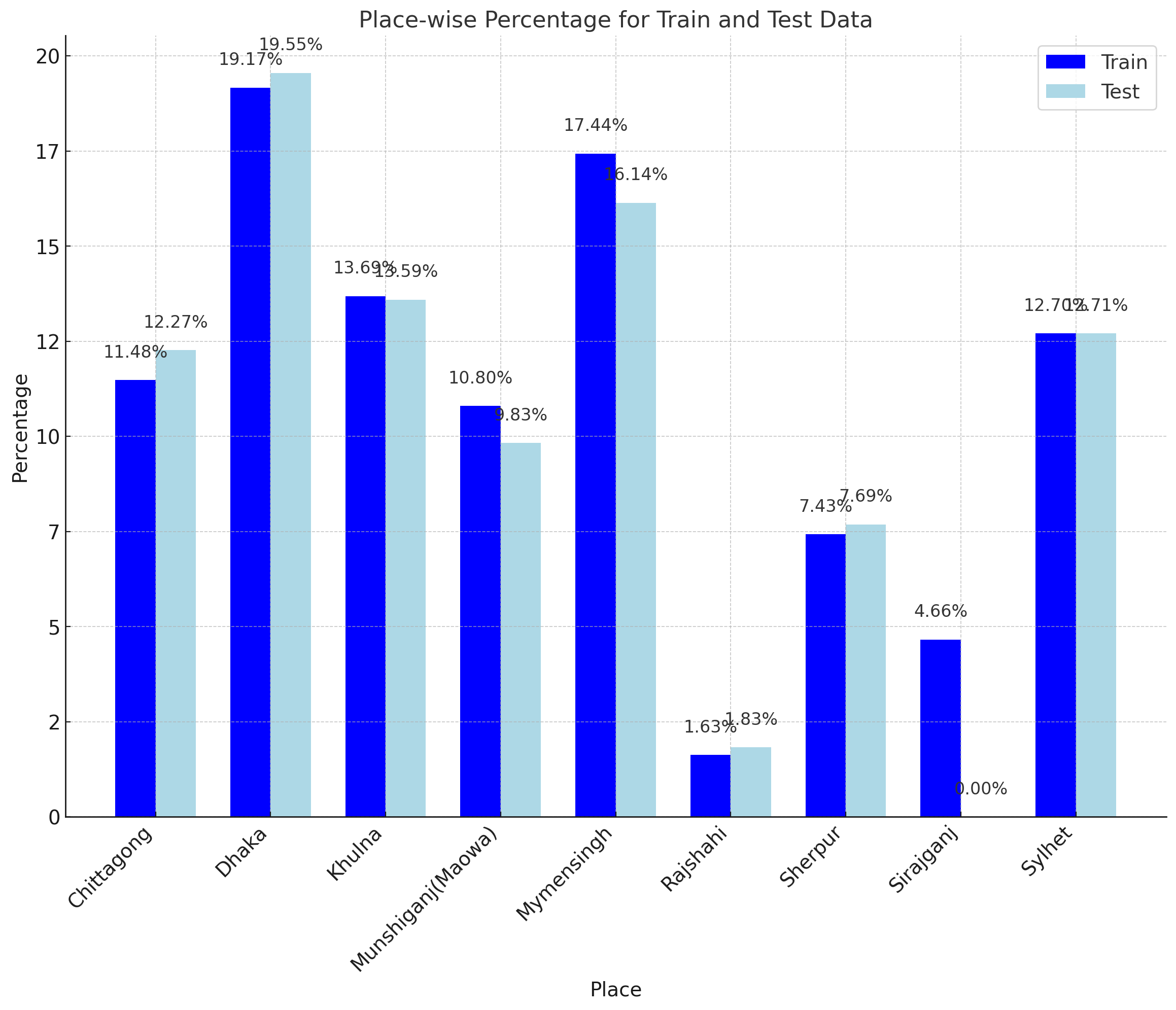}
        \caption{Location-wise train--test data distribution}
        \label{fig:train-test-coverage}
    \end{figure}

\section{Related Work}

The evolution of automatic vehicle detection (AVD) has been significantly influenced by deep learning. Initially, Convolutional Neural Networks (CNNs) were the predominant approach due to their effectiveness in handling image data. One of the pioneering models, AlexNet \cite{krizhevsky2012imagenet}, demonstrated the power of deep learning in image classification and was soon applied to AVD. This was followed by models like Faster R-CNN \cite{ren2015faster}, which introduced region proposal networks, enhancing both speed and accuracy.

Despite these advancements, CNN-based methods struggled in complex environments such as treacherous roads. Issues such as slow inference and an inability to capture global context effectively motivated the exploration of more sophisticated architectures. Single-stage detectors, such as YOLO \cite{redmon2016you}, attempted to address these problems by predicting bounding boxes and class probabilities in a single evaluation. However, such models often perform poorly in complex scenes and on small objects \cite{liu2016ssd}.

To overcome these challenges, the focus shifted to transformer-based models, initially popularized in natural language processing by Vaswani et al.~\cite{vaswani2017attention}. DETR (DEtection TRansformer)~\cite{Carion2020DETR} emerged as a landmark approach for object detection. Unlike traditional CNNs, DETR utilizes a transformer encoder-decoder architecture to predict object sets directly, capturing the global context of an image. This method eliminates the need for hand-crafted components like anchor boxes and non-maximum suppression, simplifying the detection process and improving accuracy, particularly in cluttered scenes \cite{zhu2020deformable}.

The application of DETR in AVD is still in its nascent stage. Most studies, such as those by Zhu et al.~\cite{zhu2020deformable} and Carion et al.~\cite{Carion2020DETR}, have focused on general object detection and segmentation. However, recent research indicates that integrating DETR into AVD can leverage its robust set prediction capabilities and global context understanding \cite{gao2021road}. Enhancements like Collaborative Hybrid Assignments Training (Co-DETR) show promise in improving detection accuracy and efficiency in challenging driving environments, highlighting the potential of transformer-based models in AVD.

    \begin{figure}[!htb]
        \centering
        \includegraphics[width=\linewidth]{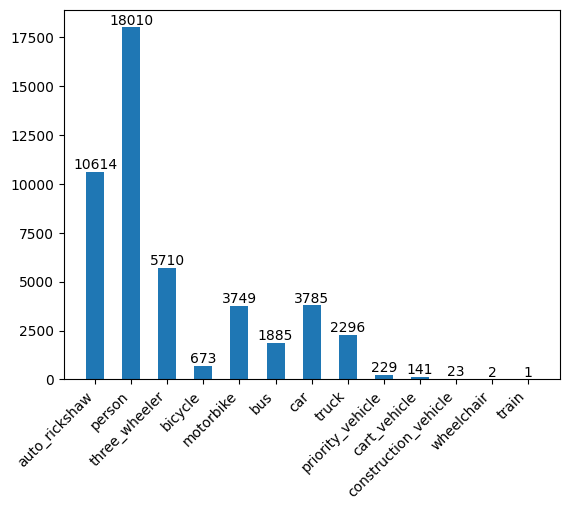}
        \caption{Class Distribution of BadODD Dataset}
        \label{fig:class-dist}
    \end{figure}

\section{Dataset Description}
    
    \subsection{Coverage of Bangladesh Districts}
        The dataset, named BadODD, covers nine districts in Bangladesh: \textbf{Sylhet, Dhaka, Rajshahi, Mymensingh, Munshiganj (Maowa), Chittagong, Sirajganj, Sherpur, and Khulna} (Fig.~\ref{fig:area-coverage}). It includes various road scenarios, such as urban and rural areas, highways, and expressways, providing a diverse representation of Bangladesh's road infrastructure. Data were collected using smartphone cameras to ensure authenticity and capture real-world driving conditions.
    
    \subsection{Image and Object Statistics}
\begin{figure*}[!htb]
            \centering
            \begin{subfigure}{0.3\linewidth}
                \centering
                \includegraphics[width=\linewidth]{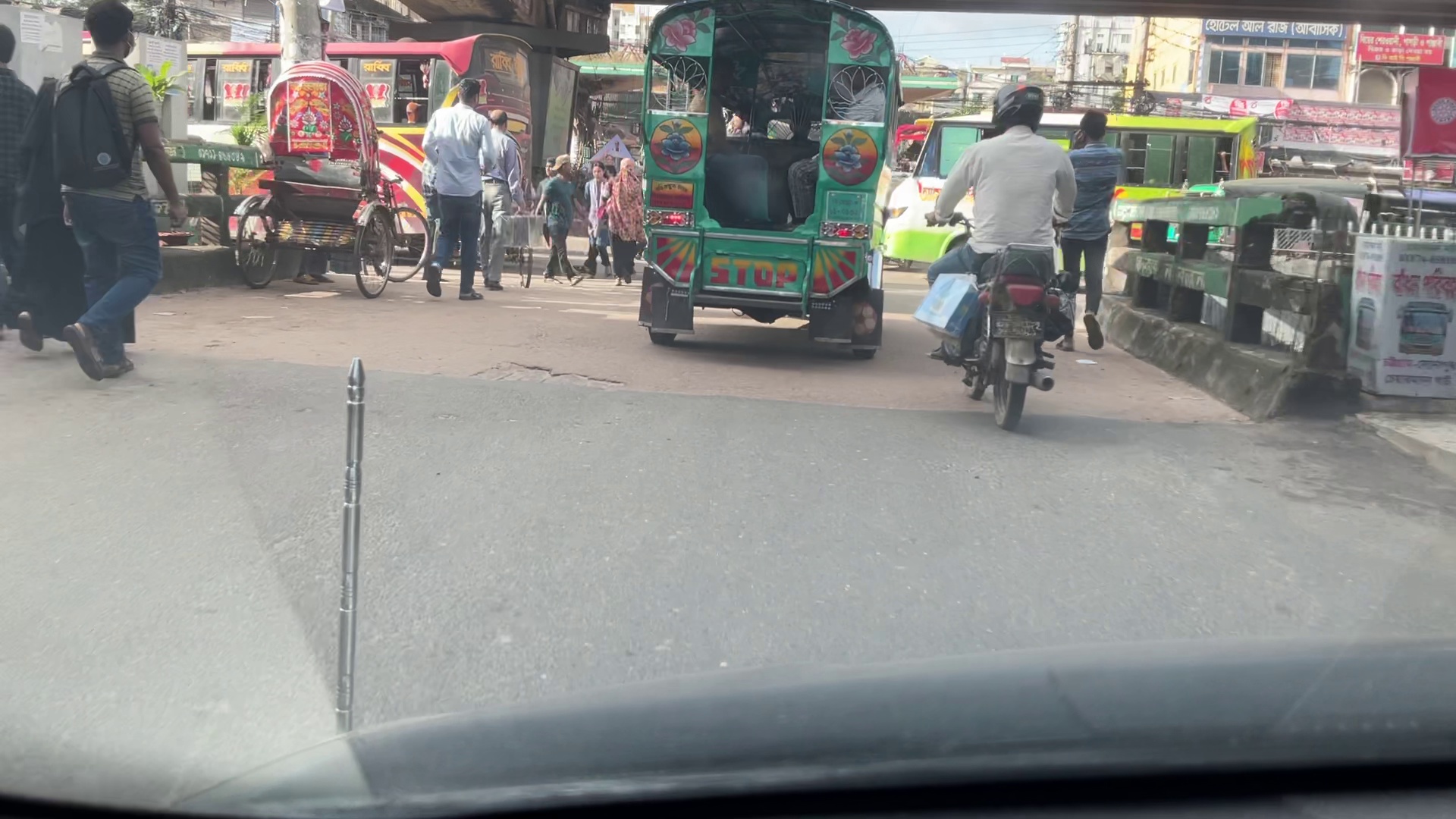}
                \caption{Day Image 1}
            \end{subfigure}
            \hfill
            \begin{subfigure}{0.3\linewidth}
                \centering
                \includegraphics[width=\linewidth]{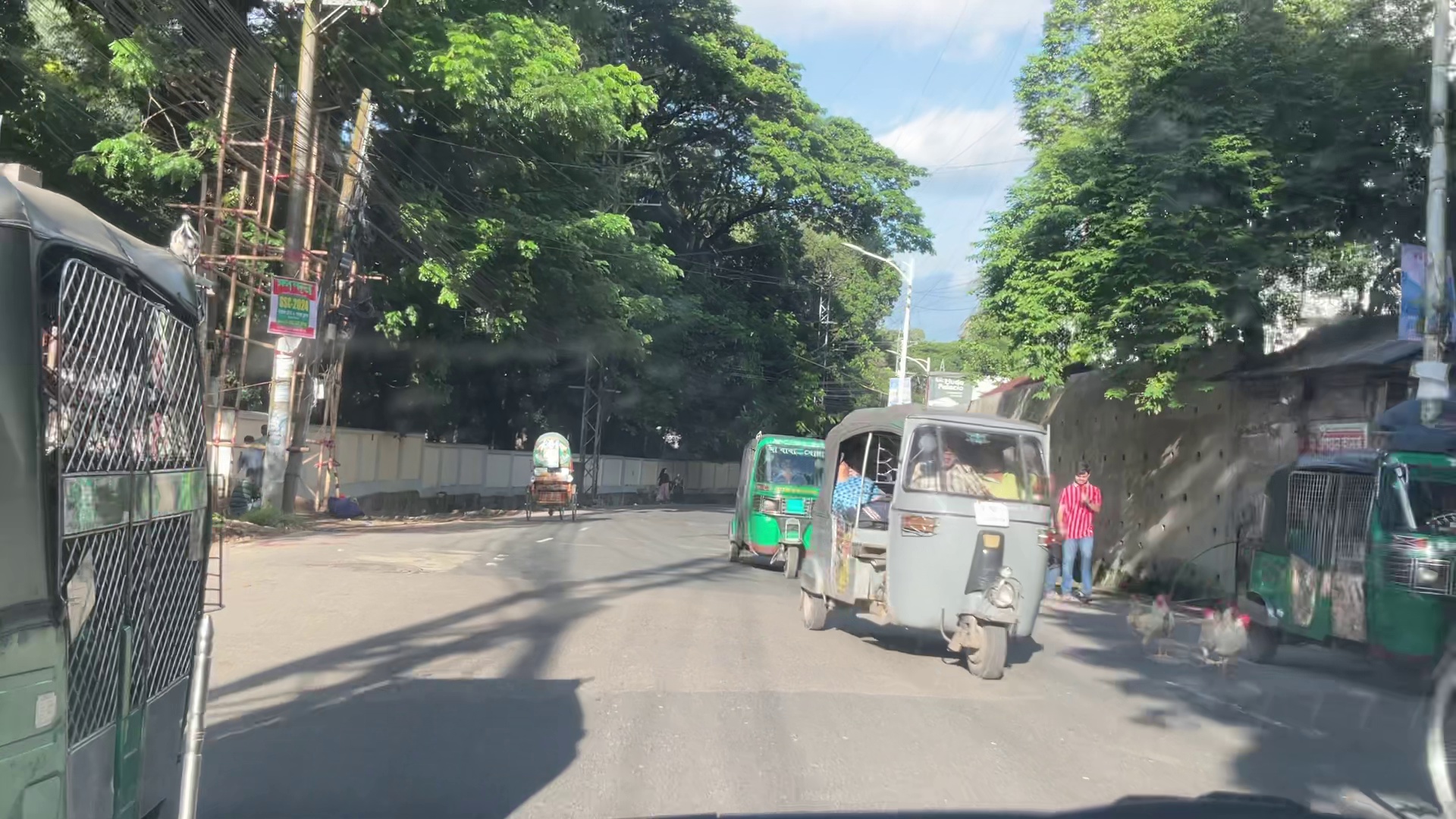}
                \caption{Day Image 2}
            \end{subfigure}
            \hfill
            \begin{subfigure}{0.3\linewidth}
                \centering
                \includegraphics[width=\linewidth]{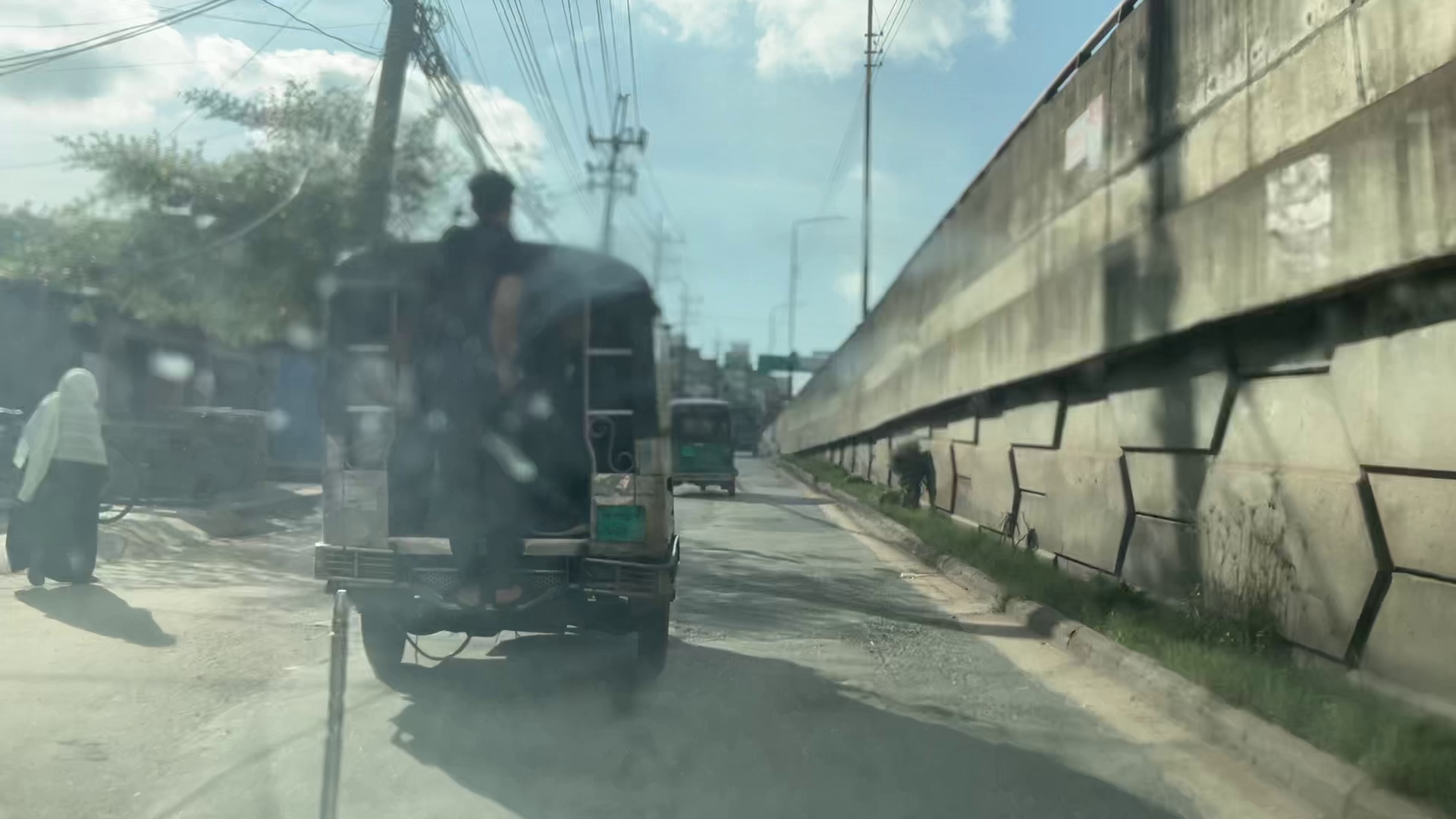}
                \caption{Day Image 3}
            \end{subfigure}
            \vskip\baselineskip
            \begin{subfigure}{0.3\linewidth}
                \centering
                \includegraphics[width=\linewidth]{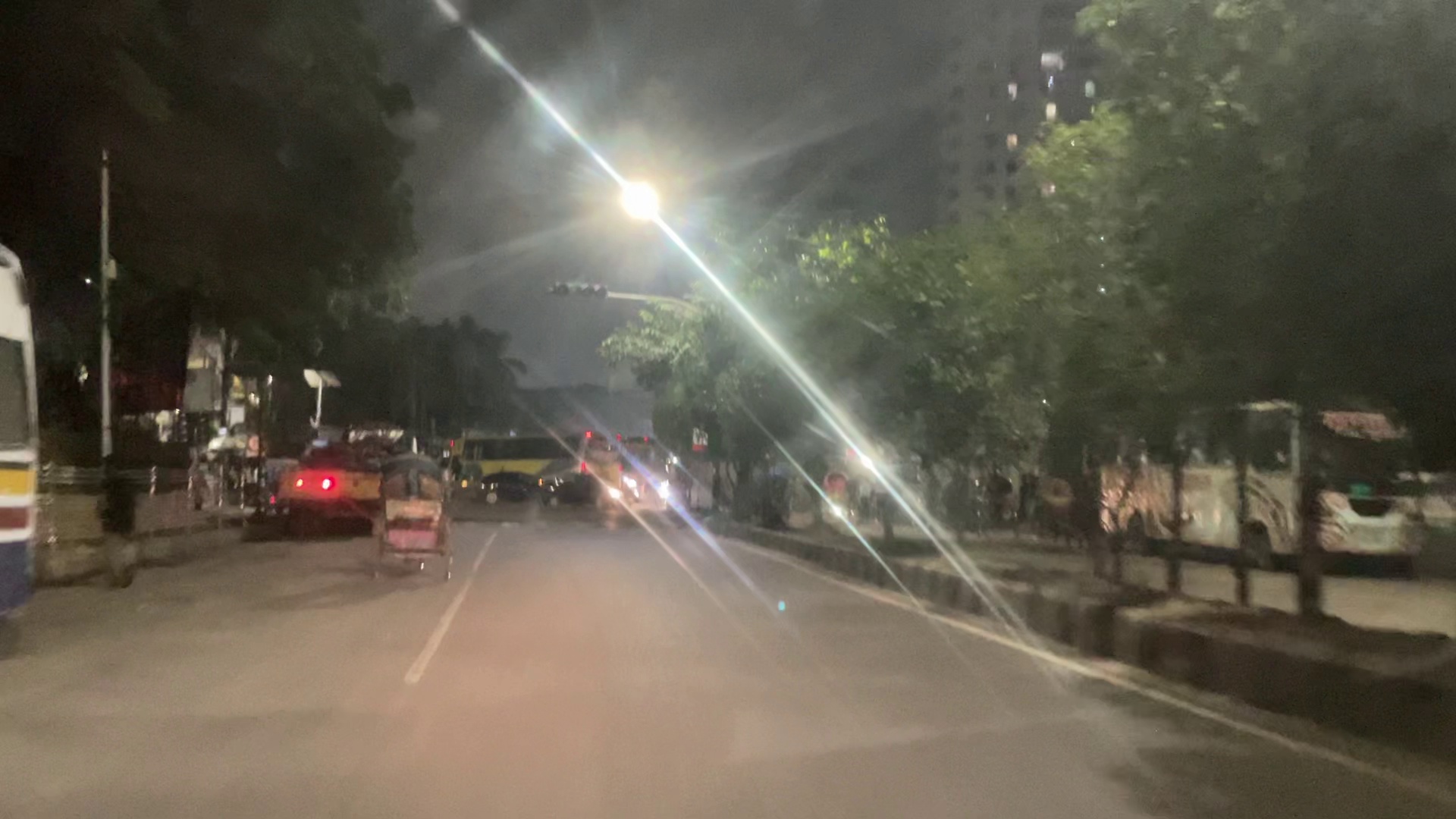}
                \caption{Night Image 1}
            \end{subfigure}
            \hfill
            \begin{subfigure}{0.3\linewidth}
                \centering
                \includegraphics[width=\linewidth]{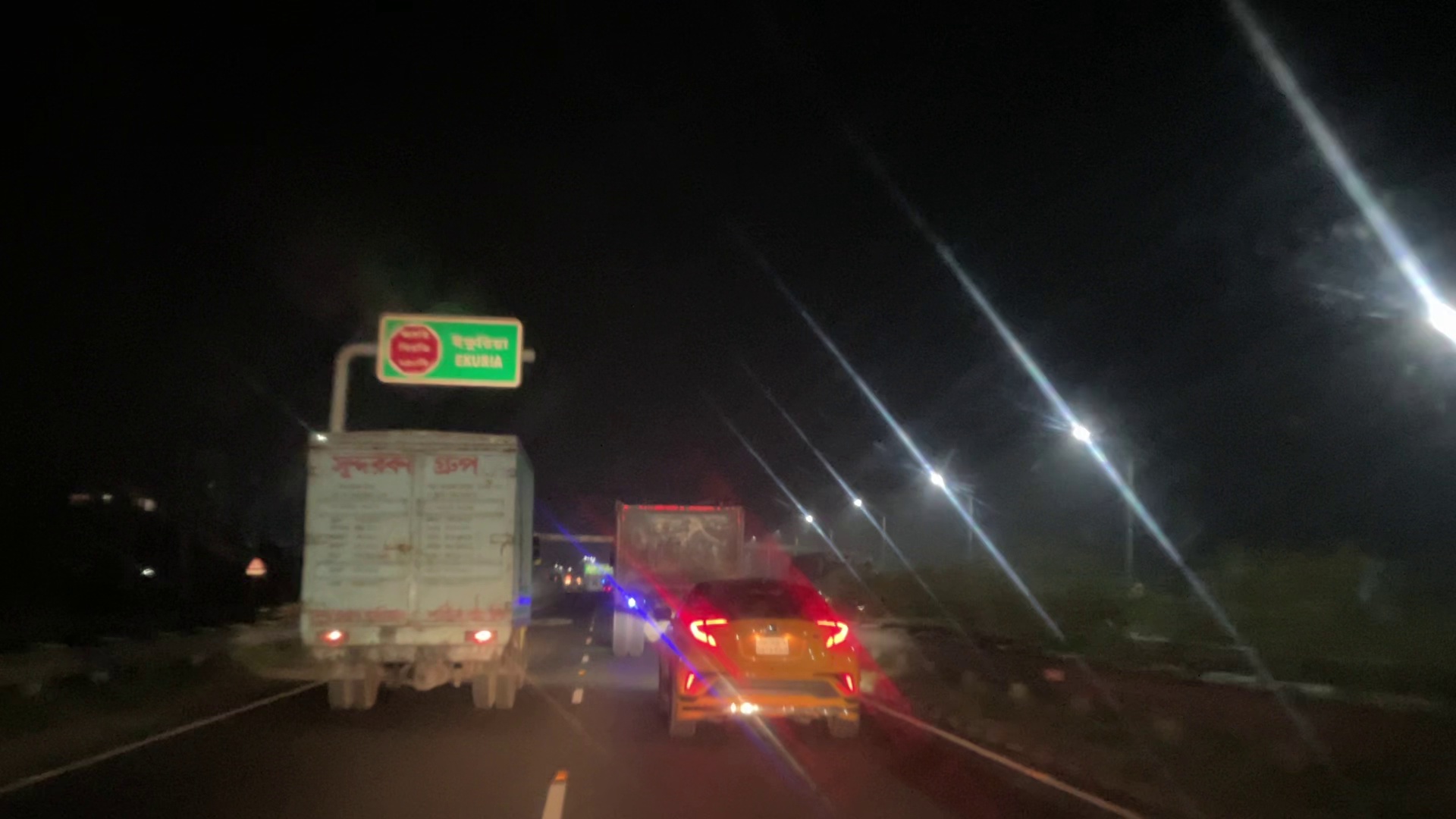}
                \caption{Night Image 2}
            \end{subfigure}
            \hfill
            \begin{subfigure}{0.3\linewidth}
                \centering
                \includegraphics[width=\linewidth]{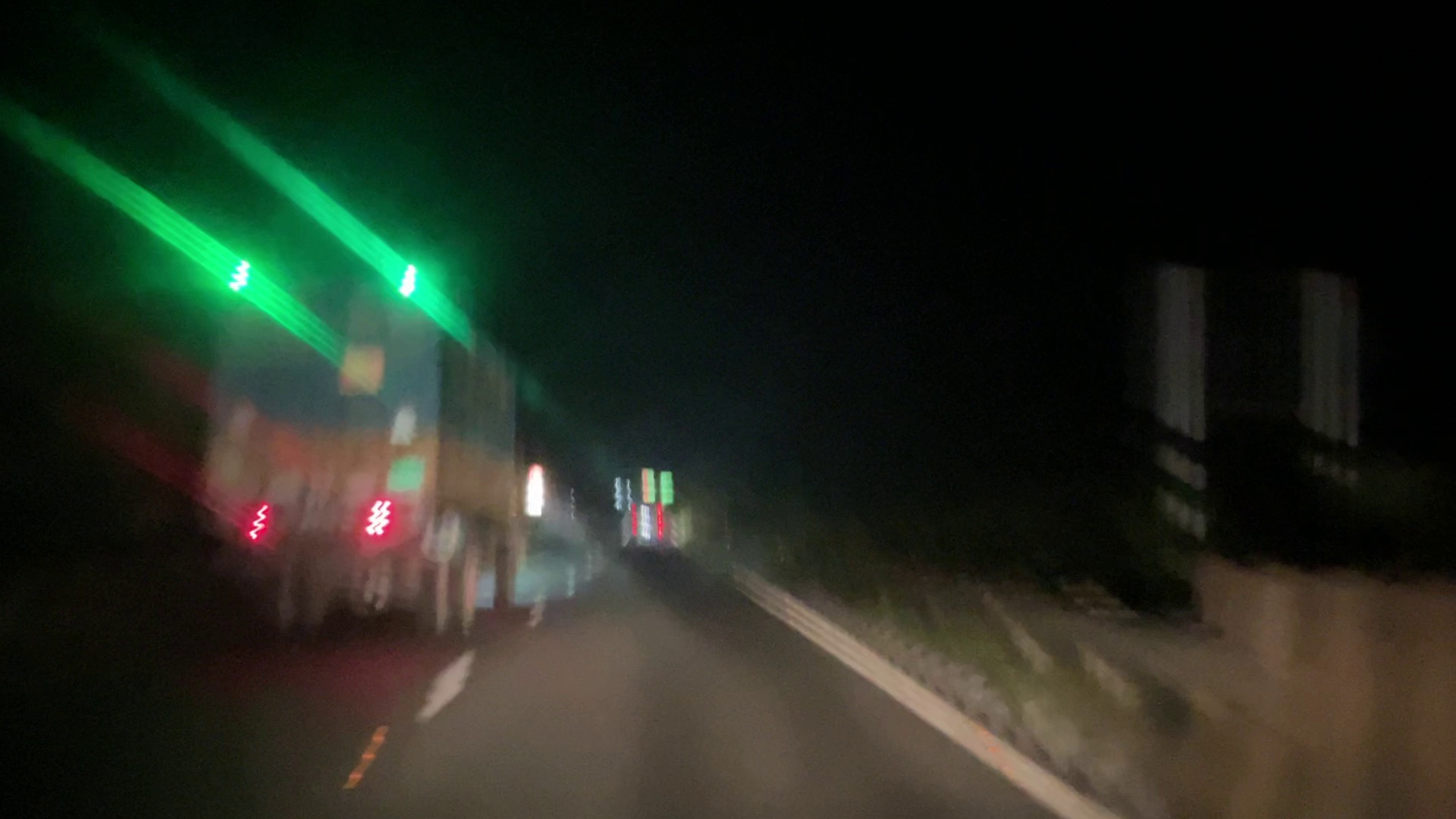}
                \caption{Night Image 3}
            \end{subfigure}
            \caption{Examples from the dataset illustrating diverse road conditions and vehicle types under different lighting scenarios. The top row (a)--(c) depicts daytime scenes, while the bottom row (d)--(f) shows nighttime scenes. This dataset is intended for studying diverse traffic patterns and vehicle behavior in Bangladesh.}
            \label{fig:diverse_driving_conditions}
        \end{figure*}

        The dataset includes a total of \textit{9,825 images} showcasing various road and driving scenarios, captured during both day and night. These are split 60:20:20 into \textit{5,896 training}, \textit{1,964 validation}, and \textit{1,965 testing} images. The class-wise train--test image distribution is shown in Fig.~\ref{fig:train-test-coverage}. There are 78,943 annotated objects across the dataset, categorized into \textit{13 distinct classes} representing vehicles commonly found under diverse driving conditions on Bangladeshi roads (Fig.~\ref{fig:diverse_driving_conditions}). Frames were selected to capture the dynamic nature of urban environments. Adaptive frame-rate sampling strategies were employed based on traffic densities, similar to other autonomous vehicle datasets such as KITTI \cite{geiger2012kitti}, nuScenes \cite{caesar2020nuscenes}, and BDD100K \cite{yu2020bdd100k}.

    \subsection{Annotation Details}
        To ensure \textbf{scalability and generalization}, classes were redefined based on vehicle characteristics rather than relying solely on local or globally recognized names. This annotation strategy is inspired by previous datasets such as nuScenes \cite{caesar2020nuscenes} and BDD100K \cite{yu2020bdd100k}, which consider environmental conditions and dataset scalability.

        Annotations were conducted using the \textbf{YOLOv5 format}, which offers simplicity and efficiency in handling diverse object classes \cite{jocher2020yolov5}. The dataset’s class distribution analysis highlights imbalances, with more prevalent classes such as \textbf{Person, Auto Rickshaw, and Three Wheeler}, while underrepresented classes include \textbf{Wheelchair, Train, and Construction Vehicle}, as shown in Fig.~\ref{fig:class-dist}. The YOLOv5 annotation format ensures efficient model training and improved performance in real-world scenarios.

     \begin{figure*}[!htb]
        \centering
        \begin{subfigure}{0.45\linewidth}
            \centering
            \includegraphics[width=\linewidth]{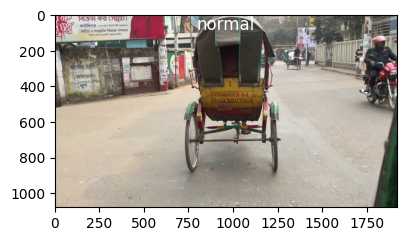}
            \caption{Raw}
        \end{subfigure}
        \hfill
        \begin{subfigure}{0.45\linewidth}
            \centering
            \includegraphics[width=\linewidth]{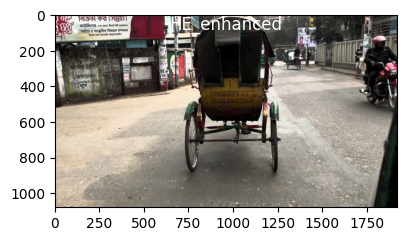}
            \caption{Histogram Equalization}
        \end{subfigure}
        \vskip\baselineskip
        \begin{subfigure}{0.45\linewidth}
            \centering
            \includegraphics[width=\linewidth]{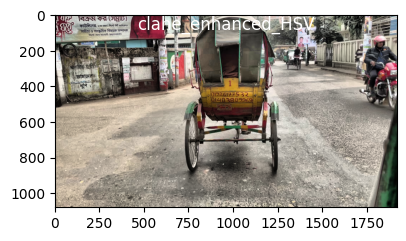}
            \caption{CLAHE}
        \end{subfigure}
        \hfill
        \begin{subfigure}{0.45\linewidth}
            \centering
            \includegraphics[width=\linewidth]{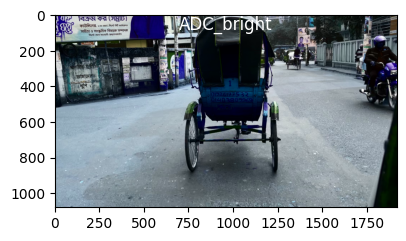}
            \caption{Gamma Correction}
        \end{subfigure}
        \caption{Comparison of image enhancement techniques applied to a raw street scene. (a) Raw image, (b) Image after Histogram Equalization, (c) Image after Contrast Limited Adaptive Histogram Equalization (CLAHE), and (d) Image after Gamma Correction. These techniques improve visual clarity and highlight different features in the dataset.}
        \label{fig:image-enhancement-techniques}
    \end{figure*}

\section{Methodology}

\subsection{Data Preprocessing}

Before feeding the data into the models for training and inference, we applied preprocessing to improve image quality and facilitate object detection. This involved several image enhancement techniques applied to the BadODD image set (Fig.~\ref{fig:image-enhancement-techniques}). These techniques improve visual quality, thereby enhancing interpretability for both humans and machine-learning models. Specifically, we employed three techniques:

\begin{itemize}
    \item \textbf{Histogram Equalization:} A method that improves image contrast by redistributing pixel intensities~\cite{Pizer1987HE}.
    \item \textbf{Contrast-Limited Adaptive Histogram Equalization (CLAHE):} An adaptive variant of histogram equalization that limits noise amplification in low-contrast regions~\cite{Zuiderveld1994CLAHE}.
    \item \textbf{Gamma Correction:} A technique that adjusts image brightness and contrast by modifying the gamma value~\cite{Fung1999GammaCorrection}.
\end{itemize}

These preprocessing steps ensured suitable input conditions for the object detectors. Fig.~\ref{fig:image-enhancement-techniques} illustrates these techniques applied to sample images from the BadODD dataset. The enhanced images served as improved input for the detection models, contributing to more accurate and reliable results.

\begin{figure*}[!t]
    \centering
    \includegraphics[width=\linewidth]{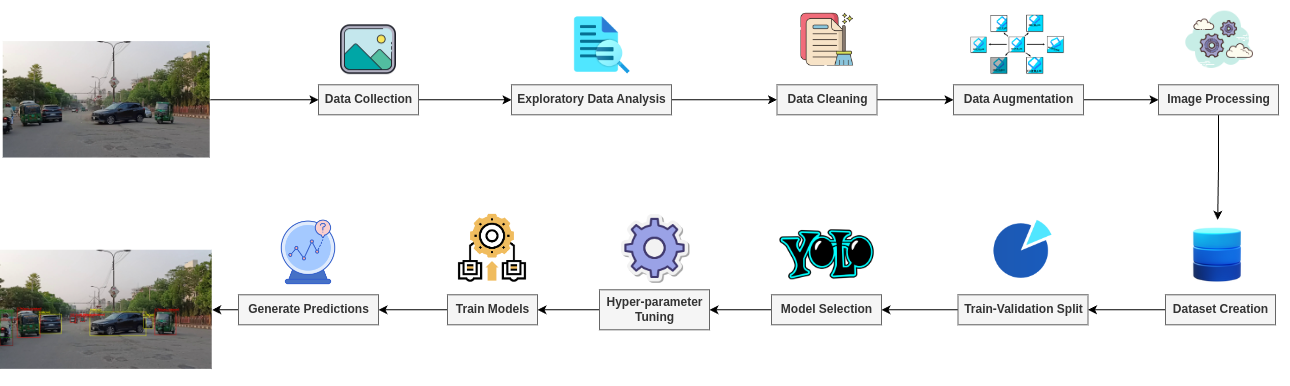}
    \caption{Workflow of the proposed methodology for automatic vehicle detection using DETR}
    \label{fig:methodology}
\end{figure*}

\subsection{Model Selection}

We initially adopted the YOLOv8m architecture for its advancements in object detection, featuring an enhanced backbone network, adaptive training strategy, and improved detection head. YOLOv8m is known for achieving superior accuracy while maintaining computational efficiency, making it a compelling choice for our study. However, our results with the current dataset did not show significant improvements.

Therefore, we opted for a transformer-based approach, known for capturing long-range dependencies and contextual information. Transformers leverage attention mechanisms, which enhance the model's understanding of global context, making them particularly suitable for complex scenes. Consequently, we explored the Co-DETR model. Co-DETR departs from traditional anchor-based methods by employing a transformer architecture and a set prediction mechanism to capture spatial relationships effectively. It predicts all object classes and their bounding boxes in a single pass, providing a robust solution for intricate scenarios.

Our investigation into both YOLOv8m and Co-DETR aimed to provide a comprehensive comparative analysis of their strengths and weaknesses in object detection. We fine-tuned these models using a dataset specifically collected for vehicle detection in Bangladesh.

    \begin{figure*}[!t]
        \centering
        \includegraphics[width=\linewidth]{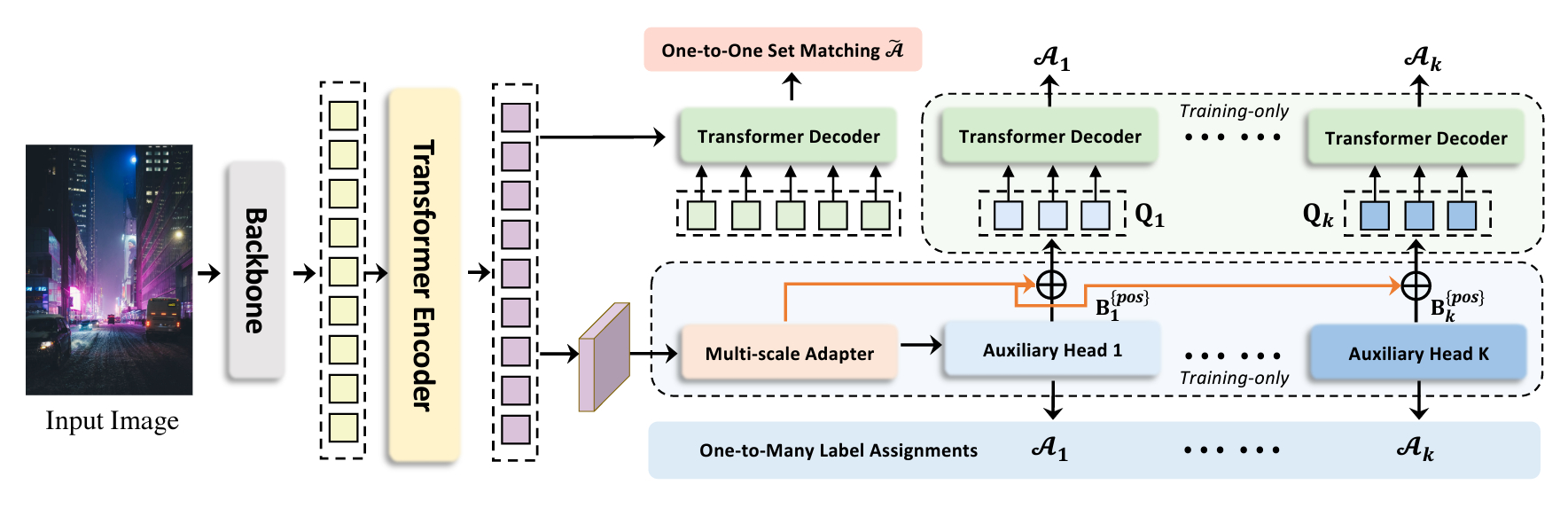}
        \caption{Framework of collaborative hybrid assignments training for DETR (Co-DETR). The auxiliary branches are discarded during inference~\cite{zong2023detrs}}
        \label{fig:co-detr-framework}
    \end{figure*}

\subsection{Model Specification}

We first fine-tuned the YOLOv8m detector as a convolutional baseline for comparison.

The training procedure involved fine-tuning the YOLOv8m model on the BadODD dataset, which was collected for detecting autonomous-driving objects on Bangladeshi roads. We configured the training process with the parameters summarized in Table~\ref{tab:yolov8-config}.

\begin{table}[ht]
    \centering
    \caption{YOLOv8m Training Configuration}
    \begin{tabular}{ll}
    \toprule
    Parameter & Value \\
    \midrule
    Model & YOLOv8m \\
    Base learning rate & 0.018 \\
    Momentum & 0.933 \\
    Cosine learning-rate schedule & Enabled \\
    Deterministic training & Enabled \\
    Random seed & 43 \\
    Evaluation metric & mAP \\
    \bottomrule
    \end{tabular}
    \label{tab:yolov8-config}
\end{table}

These parameters were chosen based on experimentation and best practices in the field of object detection for autonomous navigation tasks.

Before training the YOLOv8m model, we conducted exploratory data analysis (EDA) to better understand the dataset. We also applied image processing techniques to enhance visualization and improve object detection accuracy. Additionally, we used the predefined \textbf{validation set} of \textbf{1,964 images} (20\% of the dataset) supplied with the BadODD dataset to ensure data integrity, where images in the training set did not overlap with those in the validation set.

The YOLOv8m model underwent fine-tuning using the diverse BadODD dataset, encompassing varied driving environments across Bangladesh. However, YOLOv8m did not perform adequately on treacherous roads. To address this, we shifted to the transformer-based DETR model with Collaborative Hybrid Assignments Training (Co-DETR) for potentially improved efficiency and accuracy. Using transfer learning, we adapted the pre-trained DETR model to these challenging conditions, aiming to improve vehicle detection on difficult roadways.

Co-DETR represents a cutting-edge approach that merges transformer architecture with set prediction mechanisms, enhancing its ability to interpret complex scenes found in diverse driving conditions. By effectively capturing the global context, Co-DETR emerges as a promising alternative to the YOLOv8m baseline, particularly in environments with varied road conditions.

To optimize its performance for vehicle detection, we fine-tuned Co-DETR using a diverse dataset covering a wide range of driving scenarios in Bangladesh. We carefully adjusted key hyperparameters, including the backbone depths and query-head settings of the Swin Transformer \cite{Lin2020SwinTransformer}, to maximize accuracy and efficiency. During training, we configured the parameters summarized in Table~\ref{tab:codetr-config}.

\begin{table}[ht]
    \centering
    \caption{Co-DETR Training Configuration}
    \begin{tabular}{ll}
    \toprule
    Parameter & Value \\
    \midrule
    Model & Co-DETR \\
    Backbone & Swin Transformer \\
    Pretrained image size & 384 \\
    Embedding dimensions & 192 \\
    Depths & [2, 2, 18, 2] \\
    Number of heads & [6, 12, 24, 48] \\
    Window size & 12 \\
    MLP ratio & 4 \\
    Dropout rate & 0.3 \\
    \midrule
    Training batch size per GPU & 1 \\
    Data-loading workers & 1 \\
    Learning rate & 0.00008 \\
    Number of epochs & 10 \\
    \bottomrule
    \end{tabular}
    \label{tab:codetr-config}
\end{table}

Our objective was to harness Co-DETR's capabilities to improve object detection accuracy in the dynamic driving environments characteristic of Bangladesh. Through systematic experimentation, we evaluated its effectiveness in addressing the challenges posed by diverse driving scenarios.

\section{Experimental Setup}
All experiments were conducted on a workstation with a 16-core AMD Ryzen 9 5950X processor, 128 GB of RAM, an NVIDIA RTX 3090 GPU with 24 GB of memory, and a 2 TB storage drive, running Ubuntu 22.04 LTS. Training YOLOv8m required 1.2 hours, whereas training Co-DETR required 20 hours, both over 10 epochs. The longer training time of Co-DETR is attributed to its transformer architecture and the collaborative hybrid assignments training scheme, which involves multiple parallel auxiliary heads supervised by one-to-many label assignments. This approach improves the encoder's feature learning and the decoder's training efficiency. The auxiliary heads are discarded during inference, so they introduce no additional parameters or computational cost for the original detector.

\section{Results and Discussion}

This section evaluates the proposed approach for object detection on Bangladeshi roads using the BadODD dataset. To compare the performance of YOLOv8m and Co-DETR, we examined their mean Average Precision (mAP) scores across different training epochs, as summarized in Table~\ref{tab:map_scores}. The confidence threshold influences the number of predictions retained at evaluation time; for our evaluations, a threshold of \textbf{0.4} provided a good trade-off between prediction accuracy and detection coverage.

    \begin{table}[ht]
        \centering
        \caption{Mean Average Precision (mAP) Scores for YOLOv8m and Co-DETR Models Across Different Epochs}
        \begin{tabular}{ccc}
        \toprule
        \multirow{2}{*}{Epochs} & \multicolumn{2}{c}{mAP Scores} \\ \cmidrule(lr){2-3}
         & YOLOv8m & Co-DETR \\ 
        \midrule
        1 & 0.236 & 0.372 \\
        4 & 0.255 & 0.418 \\
        \textbf{9} & \textbf{0.295} & \textbf{0.438} \\
        \bottomrule
        \end{tabular}
        \label{tab:map_scores}
    \end{table}

Co-DETR consistently outperforms YOLOv8m at every reported epoch, demonstrating superior detection performance in diverse environments such as those found on Bangladeshi roads. It peaks at epoch 9 with an mAP of 0.438, indicating progressive refinement of detection accuracy during training.

Co-DETR achieves a peak mAP of \textit{0.438} at epoch 9, compared with \textit{0.295} for YOLOv8m at the same epoch. These results demonstrate the effectiveness of transformer-based object detection in complex road environments and support the practical deployment of Co-DETR in challenging driving scenarios.

\section{Conclusion}

This study investigated deep learning models for vehicle detection in diverse driving environments using the BadODD dataset. Co-DETR, which leverages a transformer-based architecture, outperformed the YOLOv8m baseline, achieving higher accuracy across challenging road conditions in Bangladesh. These results highlight the potential of transformer-based approaches for autonomous navigation and provide a direction for future work in vehicle detection. Further research can optimize Co-DETR for real-time applications and explore hybrid architectures that combine transformers with other deep learning paradigms.

\bibliographystyle{IEEEtran}
\bibliography{egbib}

@article{Baig2024BadODDBA,
  title={BadODD: Bangladeshi Autonomous Driving Object Detection Dataset},
  author={Mirza Nihal Baig and Rony Hajong and Mahdi Murshed Patwary and Mohammad Shahidur Rahman and Husne Ara Chowdhury},
  journal={ArXiv},
  year={2024},
  volume={abs/2401.10659},
  url={https://api.semanticscholar.org/CorpusID:267061163}
}

@inproceedings{Carion2020DETR,
  title={End-to-End Object Detection with Transformers},
  author={Nicolas Carion and Francisco Massa and Gabriel Synnaeve and Nicolas Usunier and Alexander Kirillov and Sergey Zagoruyko},
  booktitle={European Conference on Computer Vision (ECCV)},
  year={2020}
}

@inproceedings{Lin2020SwinTransformer,
  title={Swin Transformer: Hierarchical Vision Transformer using Shifted Windows},
  author={Ze Liu and Yutong Lin and Yue Cao and Han Hu and Yixuan Wei and Zheng Zhang and Stephen Lin and Baining Guo},
  booktitle={International Conference on Computer Vision (ICCV)},
  year={2021}
}

@article{Pizer1987HE,
  title={Adaptive histogram equalization and its variations},
  author={Stephen M. Pizer and E. Philip Amburn and John D. Austin and Robert Cromartie and Ari Geselowitz and Trey Greer and Bartter T. Hennessy and John H. Muller and E. Alan Gladney and Bruce C. Hanover},
  journal={Computer Vision, Graphics, and Image Processing},
  year={1987},
  volume={39},
  pages={355-368}
}

@article{Zuiderveld1994CLAHE,
  title={Contrast Limited Adaptive Histogram Equalization},
  author={Karel Zuiderveld},
  journal={Graphics Gems IV},
  year={1994},
  volume={474},
  pages={474-485}
}

@inproceedings{Fung1999GammaCorrection,
  title={Image quality enhancement using adaptive gamma correction},
  author={Wai-keung Fung and Wei Yu},
  booktitle={Proceedings 1999 International Conference on Image Processing (Cat. 99CH36348)},
  year={1999},
  volume={3},
  pages={173-176}
}

@misc{ultralytics2023yolov8,
  author       = {Jocher, Glenn and Chaurasia, Ayush and Qiu, Jing},
  title        = {{Ultralytics} {YOLO}v8},
  year         = {2023},
  howpublished = {\url{https://github.com/ultralytics/ultralytics}},
  note         = {Version 8.0.0}
}

@inproceedings{krizhevsky2012imagenet,
  title={ImageNet classification with deep convolutional neural networks},
  author={Krizhevsky, Alex and Sutskever, Ilya and Hinton, Geoffrey E},
  booktitle={Advances in neural information processing systems},
  year={2012}
}

@inproceedings{ren2015faster,
  title={Faster R-CNN: Towards real-time object detection with region proposal networks},
  author={Ren, Shaoqing and He, Kaiming and Girshick, Ross and Sun, Jian},
  booktitle={Advances in Neural Information Processing Systems (NeurIPS)},
  year={2015},
  pages={91--99}
}

@inproceedings{redmon2016you,
  title={You Only Look Once: Unified, real-time object detection},
  author={Redmon, Joseph and Divvala, Santosh and Girshick, Ross and Farhadi, Ali},
  booktitle={Proceedings of the IEEE Conference on Computer Vision and Pattern Recognition},
  year={2016}
}

@inproceedings{liu2016ssd,
  title={SSD: Single shot multibox detector},
  author={Liu, Wei and Anguelov, Dragomir and Erhan, Dumitru and Szegedy, Christian and Reed, Scott and Fu, Cheng-Yang and Berg, Alexander C},
  booktitle={European Conference on Computer Vision},
  year={2016}
}

@inproceedings{vaswani2017attention,
  title={Attention is all you need},
  author={Vaswani, Ashish and Shazeer, Noam and Parmar, Niki and Uszkoreit, Jakob and Jones, Llion and Gomez, Aidan N and Kaiser, Lukasz and Polosukhin, Illia},
  booktitle={Advances in neural information processing systems},
  year={2017}
}

@article{zhu2020deformable,
  title={Deformable DETR: Deformable transformers for end-to-end object detection},
  author={Zhu, Xizhou and Su, Weijia and Lu, Lewei and Li, Bin and Wang, Xiaogang and Dai, Jifeng},
  journal={arXiv preprint arXiv:2010.04159},
  year={2020}
}

@article{gao2021road,
  title={Road Scene Object Detection and Classification Based on DETR},
  author={Gao, Jun and Han, Song and Yang, Jian and Xue, Xiaoliang},
  journal={International Journal of Computational Intelligence Systems},
  year={2021}
}

@article{shalev2017formal,
  author    = {Shalev-Shwartz, Shai and Shammah, Shaked and Shashua, Amnon},
  title     = {On a formal model of safe and scalable self-driving cars},
  journal   = {arXiv preprint arXiv:1708.06374},
  year      = {2017}
}

@article{grigorescu2020survey,
  author    = {Grigorescu, Sorin and Trasnea, Bogdan and Cocias, Tiberiu and Macesanu, Gigel},
  title     = {A survey of deep learning techniques for autonomous driving},
  journal   = {Journal of Field Robotics},
  year      = {2020}
}

@inproceedings{viola2001rapid,
  author    = {Viola, Paul and Jones, Michael},
  title     = {Rapid object detection using a boosted cascade of simple features},
  booktitle = {Proceedings of the IEEE Conference on Computer Vision and Pattern Recognition (CVPR)},
  year      = {2001}
}

@inproceedings{dalal2005hog,
  author    = {Dalal, Navneet and Triggs, Bill},
  title     = {Histograms of oriented gradients for human detection},
  booktitle = {Proceedings of the IEEE Conference on Computer Vision and Pattern Recognition (CVPR)},
  year      = {2005}
}

@article{felzenszwalb2010object,
  author    = {Felzenszwalb, Pedro F. and Girshick, Ross B. and McAllester, David and Ramanan, Deva},
  title     = {Object detection with discriminatively trained part-based models},
  journal   = {IEEE Transactions on Pattern Analysis and Machine Intelligence (PAMI)},
  volume    = {32},
  number    = {9},
  pages     = {1627--1645},
  year      = {2010}
}

@inproceedings{zhu2019visdrone,
  author    = {Zhu, Pengfei and Wen, Longyin and Du, Dawei and Bian, Xuesong and Fan, Haibin and Hu, Qinghua and Hu, Weiming},
  title     = {VisDrone-2019: The vision meets drone object detection in image challenge results},
  booktitle = {ICCV Workshops},
  year      = {2019}
}

@inproceedings{geiger2012kitti,
  author    = {Geiger, Andreas and Lenz, Philip and Urtasun, Raquel},
  title     = {Are we ready for autonomous driving? The KITTI vision benchmark suite},
  booktitle = {Proceedings of the IEEE Conference on Computer Vision and Pattern Recognition (CVPR)},
  year      = {2012}
}

@inproceedings{caesar2020nuscenes,
  author    = {Caesar, Holger and Bankiti, Varun and Lang, Anthony H. and Vora, Sourabh and Liong, V.E. and Xu, Qiang and Krishnan, Amlan and Pan, Yu and Baldan, Giancarlo and Beijbom, Oscar},
  title     = {nuScenes: A multimodal dataset for autonomous driving},
  booktitle = {Proceedings of the IEEE Conference on Computer Vision and Pattern Recognition (CVPR)},
  year      = {2020}
}

@inproceedings{zong2023detrs,
  author    = {Zong, Zhuofan and Song, Guanglu and Liu, Yu},
  title     = {DETRs with collaborative hybrid assignments training},
  booktitle = {Proceedings of the IEEE/CVF International Conference on Computer Vision (ICCV)},
  year      = {2023},
  pages     = {6725--6735}
}

@article{yu2020bdd100k,
  author    = {Yu, Fisher and Xian, Wenqi and Chen, Yingying and Liu, Fangchen and Madhavan, Vashisht and Darrell, Trevor},
  title     = {BDD100K: A diverse driving dataset for heterogeneous multitask learning},
  journal   = {arXiv preprint arXiv:1805.04687},
  year      = {2020}
}

@misc{jocher2020yolov5,
  author    = {Jocher, Glenn and others},
  title     = {YOLOv5: Implementation of YOLO object detection in PyTorch},
  year      = {2020},
  url       = {https://github.com/ultralytics/yolov5}
}

\end{document}